\documentclass{article}
\usepackage{ijcai22}

\usepackage{times}
\usepackage{soul}
\usepackage{url}
\usepackage[hidelinks]{hyperref}
\usepackage[utf8]{inputenc}
\usepackage[small]{caption}
\usepackage{graphicx}
\usepackage{amsthm}
\usepackage{booktabs}
\usepackage{algorithm}
\usepackage{amsmath}
\usepackage{tikz}
\usepackage{bbding}
\usepackage{pifont}
\usepackage{wasysym}
\usepackage{amssymb}
\usepackage{pifont}
\usepackage{bbding}
\usepackage{pifont}
\usepackage{wasysym}
\usepackage{utfsym}
\usepackage{fontawesome}

\usepackage{algorithm}
\usepackage[noend]{algpseudocode}

\newtheorem{theorem}{Theorem}

\title{Neural PCA for Flow-Based Representation Learning}

\author{
Shen Li\footnote{Shen Li is sponsored by Google PhD fellowship 2021.}
\and
Bryan Hooi
\affiliations
Institute of Data Science, National University of Singapore
\emails
shen.li@u.nus.edu
\and
bhooi@comp.nus.edu.sg
}

\begin{document}

\maketitle

\begin{abstract}
  Of particular interest is to discover useful representations solely from observations in an unsupervised generative manner. However, the question of whether existing normalizing flows provide effective representations for downstream tasks remains mostly unanswered despite their strong ability for sample generation and density estimation. This paper investigates this problem for such a family of generative models that admits exact invertibility. We propose Neural Principal Component Analysis (Neural-PCA) that operates in full dimensionality while capturing principal components in \emph{descending} order. Without exploiting any label information, the principal components recovered store the most informative elements in their \emph{leading} dimensions and leave the negligible in the \emph{trailing} ones, allowing for clear performance improvements of $5\%$-$10\%$ in downstream tasks. Such improvements are empirically found consistent irrespective of the number of latent trailing dimensions dropped. Our work suggests that necessary inductive bias be introduced into generative modelling when representation quality is of interest.
\end{abstract}

\section{Introduction}
One of the core objectives of generative models is to deliver the promise of learning useful representations of observations in an \emph{unsupervised} manner (without exploiting any label information) \cite{bengio2013representation}. The usefulness of a representation is often measured by the representation's quality for generic downstream tasks that are discriminative in nature \cite{chapelle2009semi}\cite{dempster1977maximum}, e.g. classification tasks, and informative in describing the observations~\cite{hjelm2018learning}.



This paper investigates the usefulness of representations given by Normalizing Flows (NF) \cite{rezende2015variational}\cite{dinh2014nice}\cite{dinh2016density}\cite{kingma2018glow}\cite{NIPS2019_9183}, a family of likelihood-based models that admits exact invertibility and inference. 
We empirically find that state-of-the-art NFs yield poor representations in terms of discriminativeness and informativeness. We attribute it to a mathematical fact that an NF preserves dimensionality throughout the transform such that invertibility is guaranteed.
This, apparently, violates the manifold hypothesis that data resides in a lower dimensional manifold embedded in a full Euclidean space \cite{fefferman2016testing}. Consequently, dimensionality preservation leads to redundant representations, which undermines representations desired as shown in our empirical studies. 

To alleviate the conflict with the manifold hypothesis, we propose Neural-PCA, a flow model that operates in a full-dimensional space while capturing principal components in descending order. Without exploiting any label information, the principal components recovered in an unsupervised way store the most informative elements in leading dimensions, allowing for clear improvements in downstream tasks (e.g. classification and mutual information estimation). Empirically, we find that such improvements are consistent irrespective of the number of trailing dimensions of latent codes dropped, which acts as evidence of the manifold hypothesis. At the same time, dropping leading dimensions results in significant decline in classification accuracy and mutual information estimation, which indicates that leading dimensions capture the most dominant variations of data whereas trailing dimensions store less informative ones. Neural-PCA on one hand preserves exact invertibility, while respecting the manifold hypothesis for interpretability and better representations for downstream tasks on the other.


Specifically, a Neural-PCA can be constructed by appending the proposed PCA block to any regular normalizing flow, continuous or discrete. A PCA block is a bijection that allows for an easy inverse and an efficient evaluation of its Jacobian determinant. We further show that, to encourage principal component learning, a non-isotropic base density is a desirable choice along with Neural-PCA. Moreover, in terms of sample generation and inference, we propose an efficient approach for learning orthogonal statistics in $SO(n)$ so that Neural-PCA can be properly evaluated and inversed. Experimental results suggest that Neural-PCA captures principal components in an unsupervised yet generative manner, improving performance in downstream tasks while maintaining the ability for generating visually authentic images and for density estimation.

\section{Background}
\label{gen_inst}
A Normalizing Flow is an invertible transformation from a complex distribution to a simple probability distribution, or vice versa. Formally, let $\mathbf{z} \in \mathcal{Z} \subseteq \mathbb{R}^{n}$ be a random latent variable with a known and tractable probability density (known as base density). In the context of generative models, a normalizing flow defines an invertible differentiable function $\mathbf{f}$ such that $\mathbf{x} = \mathbf{f}(\mathbf{z})$. By using the change of variable formula \cite{papamakarios2019normalizing}, the probability density function (pdf) of $\mathbf{x}$ is given by
\begin{equation}
\label{eq:change-of-var}
p_X(\mathbf{x}) 
= p_Z({\mathbf{g(x)}}) \left\vert \det\biggl(\frac{\partial \mathbf{g}}{\partial \mathbf{x}}\biggr) \right\vert 
= p_Z(\mathbf{z}) {\left\vert \det\biggl(\frac{\partial \mathbf{f}}{\partial \mathbf{z}}\biggr) \right\vert}^{-1}
\end{equation}
where $\mathbf{g}$ is the inverse of $\mathbf{f}$. The density $p_X(\mathbf{x})$ is called a pushforward of the base density $p_Z(\mathbf{z})$ by the function $\mathbf{f}$. This movement from base density to a complex density is the \emph{generative direction}. Its inverse $\mathbf{g}$ acts in the opposite direction called \emph{normalizing direction}, which pushes a complex irregular data distribution towards a simpler or more ``normal" form. We refer readers to \cite{Kobyzev_2020} for a complete review of normalizing flows.

Throughout the entire treatment of this paper, we define a normalizing flow in the \emph{normalizing direction}: for any bijection $\mathbf{g}$, $\mathbf{x}$ denotes its input and $\mathbf{z}$ denotes the output, i.e. $\mathbf{z} = \mathbf{g}(\mathbf{x})$. Here, $\mathbf{x}$ and $\mathbf{z}$ can be, for example, 4D tensors for image data or 2D tensors for tabular data.

\section{Neural PCA}
Manifold hypothesis \cite{narayanan2010sample} is a commonly-held assumption that data are extremely concentrated on a low-dimensional manifold embedded in the ambient space with a few noisy ones residing off the manifold in various directions. The manifold is often twisted and coupled, manifesting a complex landscape. Besides merely modeling the data distribution, we expect an NF to capture major data variances into the subspace $\mathcal{Z}$ spanned by leading dimensions and leave data noise in the trailing dimensions. This inspires us to derive an operator that expands/contracts and rotates the manifold in a variance-dependent (PCA-like) manner. Experiments suggest that the proposed module will affect all the preceding transforms in terms of back-propagation, thereby leading to the desired representations. The training algorithm of Neural-PCA is summarized in Algorithm 1.

\begin{algorithm}[t]
\small
\caption{Training algorithm of Neural-PCA}
\label{alg:maxent_fewshot}
\begin{algorithmic}[1]
\Require
	Batched training data $\{X^{(1)}, ..., X^{(M)}\}$; Baseline flow $\mathbf{h_\theta}$; PCA-block $\Phi_\alpha \circ \Psi$; learnable parameters $\Theta := \{\theta, \alpha\}$.
\Ensure
	The optimal $\Theta^*$; the BatchNorm statistics $\bar{\boldsymbol\mu}, \bar{\boldsymbol\sigma}$; and $\Tilde{V}$.

\ForAll{$\text{epoch}$} %
    \ForAll{$m$ \text{in} $1, ..., M$}
		\State $Z^{(m)}, \log|\det(J_h)| \gets \mathbf{h}(X^{(m)})$;
		\State $Z^{(m)}, \log|\det(J_\Phi)| \gets  \Phi (Z^{(m)})$; \algorithmiccomment{BN Layer}
		\State $U^{(m)}, \Sigma^{(m)}, V^{(m)T}  \gets \text{SVD} (Z^{(m)})$;
        \State $Z^{(m)} \gets Z^{(m)} V^{(m)}$; \algorithmiccomment{PCA Layer}
        \State $J \gets \log p(Z^{(m)}) + \log|\det(J_g)|$;
        \State $\Theta \gets \Theta - \eta \nabla_{\Theta} J$;  \algorithmiccomment{$\eta$ is the learning rate}
	\EndFor
\EndFor
\State $\bar{\boldsymbol\mu} \gets \boldsymbol{0}, \bar{\boldsymbol\sigma} \gets \boldsymbol{0}, \overline{V} \gets \mathbf{O}$;
\ForAll{$m$ \text{in} $1, ..., M$} \algorithmiccomment{Computing training statistics}
	\State $Z^{(m)}, \log|\det(J_h)| \gets \mathbf{h}(X^{(m)})$;
	\State $Z^{(m)}, \log|\det(J_\Phi)| \gets \Phi (Z^{(m)})$;
	\State $U^{(m)}, \Sigma^{(m)}, V^{(m)T} \gets \text{SVD} (Z^{(m)})$;
	\State $\bar{\boldsymbol\mu} \gets \bar{\boldsymbol\mu} + \text{mean}(Z^{(m)})$, $\bar{\boldsymbol\sigma} \gets \bar{\boldsymbol\sigma} + \text{Var}(Z^{(m)})$;
	\State $\overline{V} \gets \overline{V} + V^{(m)}$;
\EndFor
\State $\bar{\boldsymbol\mu} \gets \bar{\boldsymbol\mu} / M$; $\bar{\boldsymbol\sigma} \gets \bar{\boldsymbol\sigma} / M$; $\overline{V} \gets \overline{V} / M$;
\State $\Tilde{V} \gets {\arg \min }_{V \in SO(n)} \|\overline{V}-V\|_{F}$; \algorithmiccomment{Mean rotation}
\end{algorithmic}
\end{algorithm}

\subsection{PCA Block}
Our proposed method, Neural-PCA, can be constructed by appending a PCA block to any regular normalizing flow $\mathbf{h}$ (defined in the normalizing direction). A PCA block consists of a batch normalization layer (BatchNorm) \cite{dinh2016density} and a PCA layer, which defines two invertible transforms: $\Phi$ and $\Psi$, respectively. 

Formally, $\Phi$ is defined as: $\mathbf{x} \xrightarrow{\Phi} \mathbf{z} := \alpha\frac{\mathbf{x} - \Tilde{\boldsymbol\mu}}{\sqrt{\Tilde{\boldsymbol\sigma}^2 + \epsilon}} + \beta$
where $\Tilde{\boldsymbol\mu}$ and $\Tilde{\boldsymbol\sigma}$ are the estimated batch mean and variance, $\alpha$ is a rescaling factor and $\beta$ is an offset. Both can be learned end-to-end. However, in an endeavor to perform principal component analysis, we fix $\beta=0$ to enforce zero mean.

After BatchNorm, we have normalized batched data $X$ as input for the PCA layer $\Psi$. 
Here, $X$ is a matrix of shape $(B, C)$, where $B$ is the batch size. Each row of $X$ represents one sample of dimensionality $C$. Then, the PCA layer performs full SVD on $X$: $X=U\Sigma V^T$, yielding a $C$-by-$C$ orthogonal rotation matrix $V$. In the forward pass, the PCA layer acts as a rotation of the data $X$ by post-multiplying $V$ such that the 1st column (dimension) contains the largest variance, followed by the 2nd, etc. Finally, the PCA layer outputs the resultant matrix $Z=XV$ of shape $(B, C)$.
In aggregate, Neural-PCA in the \emph{normalizing direction} can be constructed by composition: $\mathbf{g} := \Psi \circ \Phi \circ \mathbf{h}$.

\subsection{Choices of Base Density}
The optimization objective of Neural-PCA is in the same spirit of a regular NF, i.e. $\max_{\Theta} \mathbb{E}[\log p_{\Theta}(\mathbf{x})]$ (hereinafter w.l.o.g. we consider one sample log-likelihood and drop $\mathbb{E}[\cdot]$ to avoid notation clusters). A closed form objective can be derived by applying the change-of-variable formula~(\ref{eq:change-of-var}). Oftentimes, prior works choose an isotropic Gaussian (IG, e.g. normal distribution) as the base density for simplicity. 
In contrast, we show that this choice discourages the ordering of principal components. 
Instead, we assign different Gaussian variances to different dimensions. 
This induces a non-isotropic Gaussian as base density.
Without loss of generality, each dimension $i$ is assigned with a Gaussian variance $\sigma_i$, for $i=1,...,n$ with the relation $\sigma_1 \ge \sigma_2 \ge \cdots \ge \sigma_n > 0$ (equality holds when IG is employed).
Then, the optimization objective (denoted by $\mathcal{J}(\Theta; \sigma_1, ..., \sigma_n)$) can be written as
\begin{equation}
\begin{split}
\max_{\Theta} \mathcal{J}(\Theta; \sigma_1, ..., \sigma_n) := \log p_\Theta(\mathbf{x}) \\
= -\frac{1}{2} \sum_{i=1}^{n} \frac{z_{i}^{2}}{\sigma_{i}^{2}}-\sum_{i=1}^{n} \log \sigma_{i}-\frac{n}{2} \log 2 \pi+\log \left\vert \det\biggl(\frac{\partial \mathbf{g}_{\Theta}}{\partial \mathbf{x}}\biggr) \right\vert
\end{split}
\end{equation}
where $\mathbf{z}=\mathbf{g}_{\mathbf{\Theta}}(\mathbf{x})$ and $\Theta$ is the learnable parameters of $\mathbf{g}$.
Note that $\sigma_i$'s are predefined hyperparameters and that the optimization is performed over $\Theta$ only. Then, the objective can be simplified as
\begin{equation}
\label{eq:opt_obj}
\begin{split}
\max_{\Theta} \mathcal{J}(\Theta; \sigma_1, ..., \sigma_n) \\
=-\frac{1}{2} \sum_{i=1}^{n} \frac{z_{i}^{2}}{\sigma_{i}^{2}}-\frac{n}{2} \log 2 \pi+\log \left\vert \det\biggl(\frac{\partial \mathbf{g}_{\Theta}}{\partial \mathbf{x}}\biggr) \right\vert
\end{split}
\end{equation}

\begin{figure}
\centering
\includegraphics[width=0.45\textwidth]{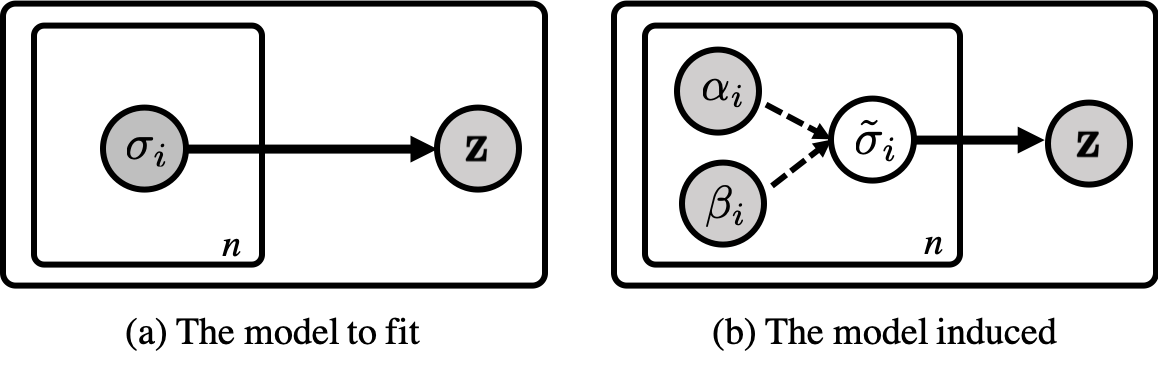}
\caption{\footnotesize Graphical models to fit and induced.}
\label{fig:model_illustr}
\end{figure}

Alternative to such a hard assignment of Gaussian variance $\sigma_i$'s, a hierarchical model shown in Figure~\ref{fig:model_illustr}(b) can relax the statistical dependency.
In experiments, we find that using the model illustrated in Figure~\ref{fig:model_illustr}(a) yields sufficiently good performance. The satisfactory performance can be explained by Theorem 1, which reveals its relation to the hierarchical Bayesian generative model with $\Tilde{\sigma}_i$'s as random variables obeying uniform distributions, i.e. $\Tilde{\sigma}_i \sim \mathcal{U}[\alpha_i, \beta_i]$.

\begin{theorem}
\label{thm:iflow_id}
Maximizing $\mathcal{J}(\Theta; \sigma_1, ..., \sigma_n)$ with respect to $\Theta$ is equivalent to maximizing a lower bound of the log-likelihood of a hierarchical Bayesian generative model with $\Tilde{\sigma}_i$'s as random variables obeying uniform distributions $\mathcal{U}[\alpha_i, \beta_i]$, where $\alpha_i$ and $\beta_i$ are subject to $\sigma_i^2=\alpha_i\beta_i$, $0<\alpha_i<\beta_i\le 1$ for all $i$'s; that is, 
\begin{equation}
\begin{split}
\max_{\Theta}  \mathcal{J}(\Theta; \sigma_1, ..., \sigma_n) \\
\le \log p(\mathbf{z} | \alpha_i, \beta_i, i=1,..., n) + \log \left|\operatorname{det}\left(\frac{\partial \mathbf{g}_{\Theta}}{\partial \mathbf{x}}\right)\right| 
\end{split}
\end{equation}
\end{theorem}

\vspace{-2mm}
\begin{proof}[Proof]
A full proof can be found in Appendix A.
\end{proof}




\paragraph{Remark 1.} Suppose all the uniform distributions have equal variance, which indicates that $\beta_i-\alpha_i=\tau$ ($\tau$ is a constant), for all $i$'s. Note that $\sigma_i = \sqrt{\alpha_i \beta_i}$. Then these two equations yield that $\sigma_i = \sqrt{\alpha_i^2+\tau \alpha_i}$, and further that $\alpha_i$ is an increasing function of $\sigma_i$ for $\sigma_i > 0$ and so is $\beta_i$. This implies that the predefined hyperparameters $\sigma_i$'s with the ordering $\sigma_1 \ge \sigma_2 \ge \cdots \ge \sigma_n > 0$ lead to the maximization of the log-likelihood lower bound of the hierarchical Bayesian generative models with $\Tilde\sigma_i$'s locating at a non-decreasing sequence of intervals $[\alpha_i, \beta_i]$, separately for all $i$'s. As shown in Figure~\ref{fig:model_illustr}, the sub-figure (a) illustrates the graphical model to fit practically and (b) demonstrates the model that (a) induces by maximizing its log-likelihood lowerbound.

\paragraph{Remark 2.} Theorem 1 suggests that $\mathcal{J}(\Theta; \sigma_1, ..., \sigma_n)$ is a lower bound of log-likelihood of the hierarchical Bayesian model induced (see Figure~\ref{fig:model_illustr}(b)) when $\beta_i \le 1$. In experiments, to achieve this theoretical bound, we enforce all $\sigma_i$'s within the open interval $(0, 1)$ evenly spaced in descending order. Since data intrinsic dimension is inaccessible and cannot be known \emph{a priori} in most cases, such an ``even-spacing" assignment is a reasonable choice for principal component learning.

\paragraph{Remark 3.} Evaluating the log-determinant term in Eq. (\ref{eq:opt_obj}) is prohibitively expensive. We propose an efficient training procedure of Neural-PCA, which requires a gradient-stopping trick. The specific treatment is relegated to Appendix B. 

\subsection{Learning Mean Rotation as Training Statistics}
In terms of sampling and inference, a regular normalizing flow with BatchNorm layers utilizes exponential moving averages of mean and variance as training statistics when the forward or inverse operation proceeds in the test mode. However, such a moving-averaging strategy cannot be directly applied to the PCA layer in Neural-PCA, since all $V$'s are in the special orthogonal group $SO(n)$ and arithmetic averaging is not closed in $SO(n)$. 

\begin{theorem}
\label{thm:proj}
For a finite positive integer $M < \infty$, the mean rotation $\mathfrak{M}(V^{(1)}, ..., V^{(M)})$ of $V^{(1)}, ..., V^{(M)} \in SO(n)$ is the orthogonal projection of the arithmetic mean of all rotation matrices $\overline{V} = \sum_{m=1}^M {V^{(m)}} / M$ in the linear space $\mathbb{R}^n$ onto the special orthogonal group $SO(n)$.
\end{theorem}
\vspace{-2mm}
\begin{proof}[Proof]
The proof is relegated to Appendix C.
\end{proof}


Fortunately, according to Theorem~\ref{thm:proj}, we can leverage the relation between the mean rotation and the usual arithmetic mean, which allows for learning training statistics in a PCA block (including the mean and variance of BatchNorm and the rotation matrix of PCA layer):

In the last training epoch, we freeze all learnable parameters $\Theta$ of Neural-PCA, and perform an extra pass over the training data to obtain arithmetic means of statistics in BatchNorm (i.e. $\Bar{\boldsymbol\mu}$ and $\Bar{\boldsymbol\sigma}$) and the arithmetic mean of $V^{(m)}$ for all $m=1,...,M$, where $M$ is the total number of batches in the last training epoch. The arithmetic means of the statistics, $\Bar{\boldsymbol\mu}$ and $\Bar{\boldsymbol\sigma}$, are used in place of the exponential moving averaged ones for the BatchNorm layer. As for the PCA layer, by virtue of Theorem~\ref{thm:proj}, we solve an additional optimization problem which finds a matrix $\Tilde{V}$ that is closest to the arithmetic mean $\overline{V}$ in the $SO(n)$ group: $\Tilde{V} = {\arg \min }_{V \in SO(n)} \|\overline{V}-V\|_{F}$,
where $\overline{V} = \sum_i {V^{(i)}/M}$ and $\Tilde{V}$ denotes the mean rotation $\mathfrak{M}(V^{(1)}, \cdots, V^{(M)})$. 
We show that this optimization problem has a closed form solution: $\Tilde{V}=Q P^T$ with the proof relegated to Appendix D. Here, $Q$ and $P$ are left- and right- singular vectors of $\overline{V}$.
Note that now the rotation matrix $\Tilde{V}$ does not depend on any test batch statistics when Neural-PCA is evaluated in the test mode. Hence, the PCA layer's forward (or inverse) operation can proceed by post-multiplying $\Tilde{V}$ (or $\Tilde{V}^T$) with zero log-determinant of Jacobian matrix. See Algorithm 1 for the entire training procedure.

\subsection{Measuring the Usefulness of Representations}
Given a flow-based representation $\mathbf{z} = \mathbf{g}(\mathbf{x}) \in \mathbb{R}^n$, we consider its corrupted representations with the leading or trailing $\kappa$ dimensions removed, denoted by $\mathbf{z}_{\downarrow \kappa^+}$ and $\mathbf{z}_{\downarrow \kappa^-}$, respectively, for $\kappa = 0, ..., n-1$. Note that $\mathbf{z}_{\downarrow 0^+} = \mathbf{z}_{\downarrow 0^-} = \mathbf{z} := \mathbf{z}_{\downarrow 0}$ which denotes the original representation $\mathbf{z}$ with zero dimensions removed. Since Neural-PCA aligns dimensions in descending order, removing trailing dimensions amounts to retaining leading principal components. We are interested in the usefulness of corrupted representations in terms of discriminativeness and informativeness, which can be measured by the following downstream metrics:
\vspace{-1mm}
\paragraph{Nonlinear classification accuracy.}  For each $\kappa$, we train a classifier using $\mathbf{z}_{\downarrow \kappa^+}$ or $\mathbf{z}_{\downarrow \kappa^-}$ of the training split of a dataset, choose the best model using the validation split and finally evaluate classification accuracy using the test split. For evaluation in nonlinear classification, we implement an MLP with one hidden layer (200 units) in the following network architecture: $(n-\kappa)$ $\mathrm{-200-ReLU-Dropout(0.2)-n_C}$, where ($n-\kappa$) is the number of remaining dimensions and $\mathrm{n_C}$ is the number of classes of a dataset in question. Note that such a narrow architecture (with 200 units) limits the expressiveness of the classifier in order to better show the class separability of the learned representations.

\vspace{-1mm}
\paragraph{Linear classification accuracy.} In similar fashion, for evaluation in linear classification, the classifier is replaced by a support vector machine (SVM) with linear kernel.

\vspace{-1mm}
\paragraph{Mutual information estimate.} Mutual information (MI) measures the statistical dependencies between two random variables. Going a step further, recent work \cite{tsai2020neural} proposed neural methods for point-wise dependency (PD) estimation, characterizing the instance-level dependency between a pair of events taken by random variables. Unlike prior methods \cite{belghazi2018mine}\cite{poole2019variational}, this fine-grained characterization is suitable for deterministic normalizing flows (when an instance $\mathbf{x}$ is given, $\mathbf{z}_{\downarrow \kappa^+}$ or $\mathbf{z}_{\downarrow \kappa^-}$ is completely determined, but not the other way around for $\kappa\neq 0$). To estimate MI between $\mathbf{x}$ and $\mathbf{z}_\kappa$ ($\kappa\neq0$), we employ the density-ratio fitting method \cite{tsai2020neural}, of which the objective is given by
\begin{equation}
\sup _{\varphi} \mathbb{E}_{P_{X, Z_{\downarrow\kappa}}}\left[\hat{r}_{\varphi}(\mathbf{x}, \mathbf{z}_{\downarrow\kappa})\right]-\frac{1}{2} \mathbb{E}_{P_{X} P_{Z_{\downarrow\kappa}}}\left[\hat{r}_{\varphi}^{2}(\mathbf{x}, \mathbf{z}_{\downarrow\kappa})\right]
\end{equation}
Here, we instantiate the density-ratio estimator as the inner-product in the embedding space followed by softplus activation to ensure positiveness, i.e. $\hat{r}_\varphi(\mathbf{x}, \mathbf{z}_{\downarrow\kappa}) := 1+\texttt{softplus} (\langle \phi(\mathbf{f}(\mathbf{x})), \phi(\mathbf{z}_{\downarrow\kappa}) \rangle)$, where $\mathbf{f}: \mathbb{R}^n \mapsto \mathbb{R}^{n-\kappa}$ is a feature map, and $\phi: \mathbb{R}^{n-\kappa} \mapsto \mathbb{R}^{d_e}$ is a nonlinear mapping with $d_e$ the dimensionality of the embedding space. Both $\mathbf{f}$ and $\phi$ are parameterized by neural networks.
Then, mutual information can be evaluated by taking the expectation of the estimated log-density-ratio: $I(X ; Z_{\downarrow\kappa}) \approx \mathbb{E}_{P_{X, Z_{\downarrow\kappa}}}\left[\log \hat{r}_{\varphi}(\mathbf{x}, \mathbf{z}_{\downarrow\kappa})\right]$.
Mathematically, higher MI, $I(X ; Z_{\downarrow\kappa})$, implies lower conditional entropy $H(X|Z_{\downarrow \kappa})$, suggesting that partial latent representation (corrupted) can be used to infer the original data $\mathbf{x}$.

\begin{figure*}[t]
\centering
\includegraphics[width=0.95\textwidth]{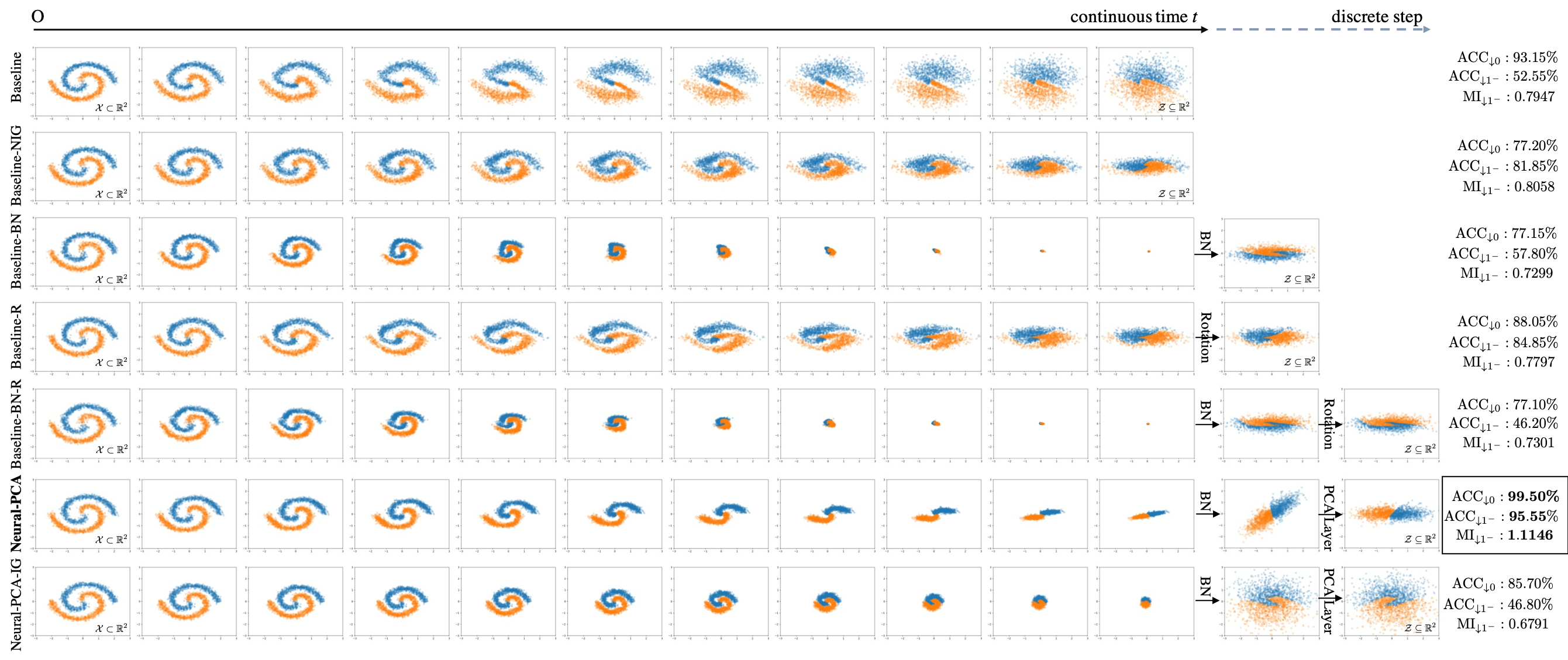}
\caption{\small Evolution of latents through time on 2D Two-Spiral in different model variants. The first column is the target density. Here, $\text{ACC}_{\downarrow 0}$ denotes test classification accuracy with full latent codes, $\text{ACC}_{\downarrow 1^-}$ and $\text{MI}_{\downarrow 1^-}$ denote test accuracy and mutual information, respectively with the last dimension removed (that is, projecting data to the horizontal axis). Neural-PCA outperforms others considerably in classification accuracy and mutual information estimate by learning linearly-separable latent representations, acting in a way of unrolling the coupled spiral.}
\label{fig:twospirals2d}
\end{figure*}

\begin{table}
\centering
\scriptsize
\begin{tabular}{ccccc}\\\toprule
\midrule
Model Name & \#Extra Params & BatchNorm & PCA Layer & Base \\\midrule
Baseline & $0$ & $-$ & $-$ & IG \\  \midrule
Baseline-NIG & $0$ & $-$ & $-$ & NIG \\  \midrule
Baseline-BN & $\mathcal{O}(n)$ & \checkmark & $-$ &  NIG \\  \midrule
Baseline-R & $\mathcal{O}(n^2)$ & $-$ & $-$ & NIG \\
\midrule
Baseline-BN-R & $\mathcal{O}(n^2)$ & \checkmark & $-$ & NIG \\    \midrule
Neural-PCA-IG & $\mathcal{O}(n)$ & \checkmark & \checkmark & IG \\  \midrule
Neural-PCA & $\mathcal{O}(n)$ & \checkmark & \checkmark & NIG \\  \bottomrule
\end{tabular}
\caption{Model variants for comparison. IG denotes isotropic Gaussian, and NIG denotes non-isotropic Gaussian. Base denotes Base Density. Here, $n$ denotes the full dimensionality.}
\label{tab:model_variants}
\end{table} 

\section{Related Work}
Finding a good rotation matrix for ICA or PCA has been explored in literature. \cite{meng2020gaussianization} proposed a trainable rotation layer, whereby the rotation matrix is parameterized with multiple trainable Householder reflections. Specifically, an $n\times n$ orthogonal matrix $R$ is represented as a product of $n$ Householder reflections: $R=H_1 H_2 \cdots H_n$, where $H_i = I - 2\mathbf{v}_i \mathbf{v}_i^T/\lVert \mathbf{v}_i \rVert_2^2$ for any vector $\mathbf{v}_i \in \mathbb{R}^n$, $i=1, ..., n$. 

However, full parameterization of a rotation matrix costs  $\mathcal{O}(n^2)$ space, which precludes its application for high dimensional data. A potential remedy is to utilize patch-based parameterization of rotation matrices \cite{meng2020gaussianization}. This amounts to learning rotation in each subspace separately, as the resulting rotation matrix is block-diagonal as a whole. This significantly reduces the number of parameters overall. However, we empirically show that the trainable rotation layer is agnostic to intrinsic data variations and thus fails to learn discriminative representations. In contrast, our mean rotation learning method is parameter-free and variation-aware, which provides useful representations for downstream tasks.

\section{Experiments}
\subsection{Implementation Details}
To demonstrate the efficacy of Neural-PCA in terms of its contributing components, throughout experiments, we consider several model variants shown in Table~\ref{tab:model_variants} for comparison. A baseline model (Baseline) is a regular normalizing flow with an isotropic Gaussian $\mathcal{N}(\mathbf{0}, I)$ as its base density. ``Baseline-NIG" is a baseline model that uses a non-isotropic Gaussian as base density. This is to show the efficacy of the PCA block of Neural-PCA. ``Baseline-BN" is a baseline model with a BatchNorm layer at the end. This is to show the necessity of PCA layer in a PCA block. ``Baseline-R" is a baseline model with a trainable rotation layer. ``Neural-PCA-IG" is the same as Neural-PCA except for the choice of an isotropic Gaussian $\mathcal{N}(\mathbf{0}, I)$ as base density. 
We show in experiments that to learn useful representations, the PCA block and non-isotropy of base density are two essential building blocks complementary to one another.


\subsection{Evaluation on Toy Datasets}
For intuitive understanding, we conduct experiments on a two-dimensional toy dataset, Two-Spiral. As shown in Figure~\ref{fig:twospirals2d}, each of the two colors represents one class. However, this label information is not given during training, as the goal is to show improvements on downstream tasks with latent codes that are learned in an unsupervised manner. 


We use the state-of-the-art flow, FFJORD \cite{grathwohl2018ffjord}, as the baseline flow for modeling distribution of tabular data. FFJORD is a continuous flow that allows unrestricted architectures as compared to Neural ODE \cite{chen2018neural}.
For an instantiation of Neural-PCA, a PCA block is appended to a baseline flow. 
To encourage the latent ordering, a non-isotropic Gaussian density $\mathcal{N}(\mathbf{0}, \text{diag}(1, 0.1))$ is used as the base density. The batch size for training is set to 100 for all models. For Neural-PCA, the proposed projection method (cf. Section 3.3) is utilized to aggregate all rotation matrices computed from different batches. After training, models are evaluated on a linear classification task using the learned representations of $2000$ test points. To measure linear separability, we use SVM with linear kernel as the binary classifier. 

Ideally, as a diffeomorphism, a well-fitted flow model should unroll a spiral into a finite line segment lying in the latent space. The length of the spiral should be proportional to the largest variance and the thickness to the least. Further, to encourage axis-aligned representations, a non-isotropic Gaussian density with a diagonal covariance matrix is chosen as the base density. The ordering of the diagonals encodes our prior belief and encourages the desired unrolling behavior. This hypothesis verifies our choices for the base density, a non-isotropic Gaussian $\mathcal{N}(\mathbf{0}, \text{diag}(1, 0.1))$. As shown in Figure~\ref{fig:twospirals2d}, Neural-PCA is capable of unrolling the spiral, whereas other models contract or expand the spiral, disrespecting its intrinsic manifold. Intriguingly, this unrolling behavior results in far clearer linear separability in the latent space ($99.50\%$ accuracy). Also noting that, for Neural-PCA after PCA layer, the horizontal dimension contains the largest variation and the vertical has the least, we observe that the classification boundary is almost perpendicular to the first major axis. This suggests that these two classes can still be well-separated even after dimensionality reduction (projecting points to the horizontal axis achieves $95.55\%$ accuracy). The other methods apparently fail to do so. This visually reveals why Neural-PCA outperforms the others in learning discriminative representations of data. 
Detailed comparison and analyses can be found in Appendix E.

\begin{figure*}[t]
\centering
\includegraphics[width=\textwidth]{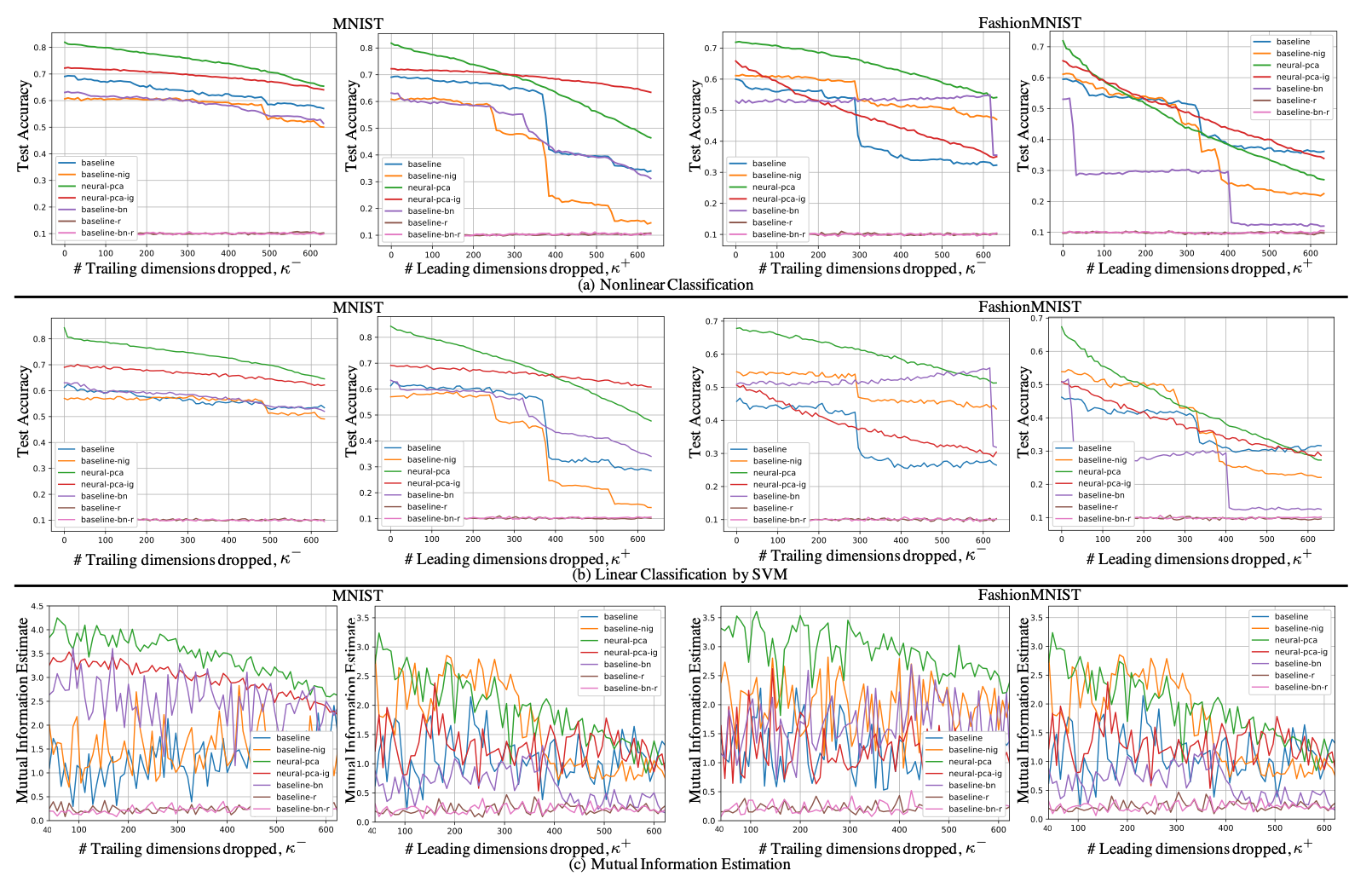}
\caption{\small Comparison of the usefulness of representations learned by different models in terms of (a) nonlinear classification, (b) linear classification by SVM and (c) mutual information estimation.}
\label{fig:exp_res}
\end{figure*}



\subsection{Evaluation on High-Dimensional Datasets}
To demonstrate the effectiveness of Neural-PCA on high-dimensional data, we conduct evaluations on image datasets: MNIST and FashionMNIST, which both give $1024 (=32\times32\times1)$-dimensional spaces. Note that the original spatial sizes of MNIST and FashionMNIST data are $(28\times28\times1)$. Preprocessing includes padding zeros on boundaries for such data. Dequantization is also performed by adding uniform noise to the images~\cite{uria2013rnade}\cite{theis2015note}.


We choose the state-of-the-art flow, RQ-NSF(C) \cite{durkan2019neural}, as the baseline flow for modeling image data. RQ-NSF(C) is an analytically invertible model based on monotonic rational-quadratic splines that improves the flexibility of coupling transforms. Similarly as in toy data, to encourage the ordering, we set $\sigma_i$'s evenly from $1.0$ down to $0.005$, respectively for $i=1,\cdots, n$. All models are trained for 100,000 iterations with Adam optimizer. The batch size is set to 1800. Learning rates are annealed from $5\times10^{-4}$ down to zero following a cosine learning policy. The number of positions $L$ is set to 1. 
Similarly in the Two-Spiral setting, mean rotations are calculated using the projection method proposed in Section 3.3.



\vspace{-1mm}
\paragraph{Nonlinear Classification.} As shown in Figure~\ref{fig:exp_res}(a), Neural-PCA outperforms all the other model variants by large margins when zero dimensions are removed as well as when a large fraction of trailing dimensions are removed ($624 > 1024/2$). 
Removing leading dimensions, on the other hand, gives rise to rapid decline in test accuracy of Neural-PCA, which suggests that the principal components recovered in our proposed unsupervised manner manifests strong discriminativeness of the data. We observe that Baseline-R and Baseline-BN-R learn poor representations for classification with test accuracy $\sim10\%$; as the number of categories is 10 for all datasets, these two models amount to random guess.

\vspace{-1mm}
\paragraph{Linear Classification by SVM.} For linear classification, we use SVM with linear kernel in place of the MLP as in nonlinear classification. As shown in Figure~\ref{fig:exp_res}(b), similar behaviors can be observed: substantial accuracy improvements of $\gtrsim 10\%$ on MNIST and FashionMNIST when zero dimensions are removed. The improvements remain clear even after removing a large proportion of trailing dimensions.

\vspace{-1mm}
\paragraph{Mutual Information Estimation.} In terms of mutual information estimation, similar and consistent behaviors of Neural-PCA as in classification metrics can be observed, as suggested in Figure 3(c).

\subsection{Comparison with Post-PCA models.}
This section demonstrates the necessity of training flow with PCA-Block end-to-end jointly. As in the toy data setting, we consider post-PCA models based on baseline features learned by Baseline, Baseline-NIG, Baseline-BN, Baseline-BN-R and Neural-PCA-IG. Our results suggest that the representations learned by these models have already been axis-aligned, and therefore post-PCA on them would not result in non-trivial rotation further. Comparison results and detailed analysis can be found in Appendix F.

\vspace{-1mm}
\subsection{Miscellaneous}
Other interesting properties of Neural-PCA, including generation quality, density estimation, mean rotation analysis and latent interpolation, are relegated to Appendix G, H and I, respectively. The implementation details for experiments can be found in Appendix J. Code is publicly available at \texttt{https://github.com/MathsShen/Neural-PCA}.

\vspace{-1mm}
\section{Conclusions}
\vspace{-1mm}
We have proposed Neural-PCA that operates in full dimensionality while capturing principal components in descending order. Consequently, the principal components recovered store the most informative elements in leading dimensions and leave the negligible in trailing ones, allowing for considerable improvements in downstream tasks. Echoing the manifold hypothesis, such improvements are empirically found consistent irrespective of the number of trailing dimensions dropped. Our work suggests that necessary inductive biases be introduced into generative modelling when representation quality is of interest. 









\section*{Acknowledgements}
We thank all anonymous reviewers for constructive suggestions on the manuscript of this paper. This work is in memoriam of Shen Li's Grandmother, Baoling Hong (1939-2021).

\bibliographystyle{named}
\bibliography{ijcai22}

\begin{thebibliography}{}

\bibitem[\protect\citeauthoryear{Belghazi \bgroup \em et al.\egroup
  }{2018}]{belghazi2018mine}
Mohamed~Ishmael Belghazi, Aristide Baratin, Sai Rajeswar, Sherjil Ozair, Yoshua
  Bengio, Aaron Courville, and R~Devon Hjelm.
\newblock Mine: mutual information neural estimation.
\newblock {\em arXiv preprint arXiv:1801.04062}, 2018.

\bibitem[\protect\citeauthoryear{Bengio \bgroup \em et al.\egroup
  }{2013}]{bengio2013representation}
Yoshua Bengio, Aaron Courville, and Pascal Vincent.
\newblock Representation learning: A review and new perspectives.
\newblock {\em IEEE transactions on pattern analysis and machine intelligence},
  35(8):1798--1828, 2013.

\bibitem[\protect\citeauthoryear{Chapelle \bgroup \em et al.\egroup
  }{2009}]{chapelle2009semi}
Olivier Chapelle, Bernhard Scholkopf, and Alexander Zien.
\newblock Semi-supervised learning (chapelle, o. et al., eds.; 2006)[book
  reviews].
\newblock {\em IEEE Transactions on Neural Networks}, 20(3):542--542, 2009.

\bibitem[\protect\citeauthoryear{Chen \bgroup \em et al.\egroup
  }{2018}]{chen2018neural}
Ricky~TQ Chen, Yulia Rubanova, Jesse Bettencourt, and David~K Duvenaud.
\newblock Neural ordinary differential equations.
\newblock In {\em Advances in neural information processing systems}, pages
  6571--6583, 2018.

\bibitem[\protect\citeauthoryear{Chen \bgroup \em et al.\egroup
  }{2019}]{NIPS2019_9183}
Ricky T.~Q. Chen, Jens Behrmann, David~K Duvenaud, and Joern-Henrik Jacobsen.
\newblock Residual flows for invertible generative modeling.
\newblock In H.~Wallach, H.~Larochelle, A.~Beygelzimer, F.~d\textquotesingle
  Alch\'{e}-Buc, E.~Fox, and R.~Garnett, editors, {\em Advances in Neural
  Information Processing Systems 32}, pages 9916--9926. Curran Associates,
  Inc., 2019.

\bibitem[\protect\citeauthoryear{Dempster \bgroup \em et al.\egroup
  }{1977}]{dempster1977maximum}
Arthur~P Dempster, Nan~M Laird, and Donald~B Rubin.
\newblock Maximum likelihood from incomplete data via the em algorithm.
\newblock {\em Journal of the Royal Statistical Society: Series B
  (Methodological)}, 39(1):1--22, 1977.

\bibitem[\protect\citeauthoryear{Dinh \bgroup \em et al.\egroup
  }{2014}]{dinh2014nice}
Laurent Dinh, David Krueger, and Yoshua Bengio.
\newblock Nice: Non-linear independent components estimation.
\newblock {\em arXiv preprint arXiv:1410.8516}, 2014.

\bibitem[\protect\citeauthoryear{Dinh \bgroup \em et al.\egroup
  }{2016}]{dinh2016density}
Laurent Dinh, Jascha Sohl-Dickstein, and Samy Bengio.
\newblock Density estimation using real nvp.
\newblock {\em International Conference on Learning Representations}, 2016.

\bibitem[\protect\citeauthoryear{Durkan \bgroup \em et al.\egroup
  }{2019}]{durkan2019neural}
Conor Durkan, Artur Bekasov, Iain Murray, and George Papamakarios.
\newblock Neural spline flows.
\newblock In {\em Advances in Neural Information Processing Systems}, pages
  7509--7520, 2019.

\bibitem[\protect\citeauthoryear{Fefferman \bgroup \em et al.\egroup
  }{2016}]{fefferman2016testing}
Charles Fefferman, Sanjoy Mitter, and Hariharan Narayanan.
\newblock Testing the manifold hypothesis.
\newblock {\em Journal of the American Mathematical Society}, 29(4):983--1049,
  2016.

\bibitem[\protect\citeauthoryear{Grathwohl \bgroup \em et al.\egroup
  }{2018}]{grathwohl2018ffjord}
Will Grathwohl, Ricky~TQ Chen, Jesse Bettencourt, Ilya Sutskever, and David
  Duvenaud.
\newblock Ffjord: Free-form continuous dynamics for scalable reversible
  generative models.
\newblock In {\em International Conference on Learning Representations}, 2018.

\bibitem[\protect\citeauthoryear{He \bgroup \em et al.\egroup
  }{2020}]{he2020momentum}
Kaiming He, Haoqi Fan, Yuxin Wu, Saining Xie, and Ross Girshick.
\newblock Momentum contrast for unsupervised visual representation learning.
\newblock In {\em Proceedings of the IEEE/CVF Conference on Computer Vision and
  Pattern Recognition}, pages 9729--9738, 2020.

\bibitem[\protect\citeauthoryear{Hjelm \bgroup \em et al.\egroup
  }{2018}]{hjelm2018learning}
R~Devon Hjelm, Alex Fedorov, Samuel Lavoie-Marchildon, Karan Grewal, Phil
  Bachman, Adam Trischler, and Yoshua Bengio.
\newblock Learning deep representations by mutual information estimation and
  maximization.
\newblock {\em arXiv preprint arXiv:1808.06670}, 2018.

\bibitem[\protect\citeauthoryear{Ionescu \bgroup \em et al.\egroup
  }{2015}]{ionescu2015matrix}
Catalin Ionescu, Orestis Vantzos, and Cristian Sminchisescu.
\newblock Matrix backpropagation for deep networks with structured layers.
\newblock In {\em Proceedings of the IEEE International Conference on Computer
  Vision}, pages 2965--2973, 2015.

\bibitem[\protect\citeauthoryear{Jaini \bgroup \em et al.\egroup
  }{2019}]{jaini2019sum}
Priyank Jaini, Kira~A Selby, and Yaoliang Yu.
\newblock Sum-of-squares polynomial flow.
\newblock In {\em International Conference on Machine Learning}, pages
  3009--3018. PMLR, 2019.

\bibitem[\protect\citeauthoryear{Kingma and Dhariwal}{2018}]{kingma2018glow}
Durk~P Kingma and Prafulla Dhariwal.
\newblock Glow: Generative flow with invertible 1x1 convolutions.
\newblock In {\em Advances in Neural Information Processing Systems}, pages
  10215--10224, 2018.

\bibitem[\protect\citeauthoryear{Kobyzev \bgroup \em et al.\egroup
  }{2020}]{Kobyzev_2020}
Ivan Kobyzev, Simon Prince, and Marcus Brubaker.
\newblock Normalizing flows: An introduction and review of current methods.
\newblock {\em IEEE Transactions on Pattern Analysis and Machine Intelligence},
  page 1–1, 2020.

\bibitem[\protect\citeauthoryear{Meng \bgroup \em et al.\egroup
  }{2020}]{meng2020gaussianization}
Chenlin Meng, Yang Song, Jiaming Song, and Stefano Ermon.
\newblock Gaussianization flows.
\newblock In {\em International Conference on Artificial Intelligence and
  Statistics}, 2020.

\bibitem[\protect\citeauthoryear{Narayanan and
  Mitter}{2010}]{narayanan2010sample}
Hariharan Narayanan and Sanjoy Mitter.
\newblock Sample complexity of testing the manifold hypothesis.
\newblock In {\em Proceedings of the 23rd International Conference on Neural
  Information Processing Systems-Volume 2}, pages 1786--1794, 2010.

\bibitem[\protect\citeauthoryear{Papamakarios \bgroup \em et al.\egroup
  }{2017}]{papamakarios2017masked}
George Papamakarios, Theo Pavlakou, and Iain Murray.
\newblock Masked autoregressive flow for density estimation.
\newblock In {\em Advances in Neural Information Processing Systems}, pages
  2338--2347, 2017.

\bibitem[\protect\citeauthoryear{Papamakarios \bgroup \em et al.\egroup
  }{2019}]{papamakarios2019normalizing}
George Papamakarios, Eric Nalisnick, Danilo~Jimenez Rezende, Shakir Mohamed,
  and Balaji Lakshminarayanan.
\newblock Normalizing flows for probabilistic modeling and inference.
\newblock {\em arXiv preprint arXiv:1912.02762}, 2019.

\bibitem[\protect\citeauthoryear{Poole \bgroup \em et al.\egroup
  }{2019}]{poole2019variational}
Ben Poole, Sherjil Ozair, Aaron Van Den~Oord, Alex Alemi, and George Tucker.
\newblock On variational bounds of mutual information.
\newblock In {\em International Conference on Machine Learning}, pages
  5171--5180. PMLR, 2019.

\bibitem[\protect\citeauthoryear{Rezende and
  Mohamed}{2015}]{rezende2015variational}
Danilo~Jimenez Rezende and Shakir Mohamed.
\newblock Variational inference with normalizing flows.
\newblock {\em International Conference on Machine Learning}, 2015.

\bibitem[\protect\citeauthoryear{Theis \bgroup \em et al.\egroup
  }{2015}]{theis2015note}
Lucas Theis, A{\"a}ron van~den Oord, and Matthias Bethge.
\newblock A note on the evaluation of generative models.
\newblock {\em arXiv preprint arXiv:1511.01844}, 2015.

\bibitem[\protect\citeauthoryear{Tsai \bgroup \em et al.\egroup
  }{2020}]{tsai2020neural}
Yao-Hung~Hubert Tsai, Han Zhao, Makoto Yamada, Louis-Philippe Morency, and
  Ruslan Salakhutdinov.
\newblock Neural methods for point-wise dependency estimation.
\newblock {\em arXiv preprint arXiv:2006.05553}, 2020.

\bibitem[\protect\citeauthoryear{Uria \bgroup \em et al.\egroup
  }{2013}]{uria2013rnade}
Benigno Uria, Iain Murray, and Hugo Larochelle.
\newblock Rnade: the real-valued neural autoregressive density-estimator.
\newblock In {\em Proceedings of the 26th International Conference on Neural
  Information Processing Systems-Volume 2}, pages 2175--2183, 2013.

\end{thebibliography}

\clearpage
\onecolumn

\renewcommand\thetable{A.\arabic{table}}
\renewcommand\thefigure{A.\arabic{figure}}
\renewcommand\theequation{A.\arabic{equation}}

\appendix

\section{Proof of Theorem 1}
\begin{proof}
Note that for any $\sigma_i^2>0$, there exist $\alpha_i$ and $\beta_i$ such that $\sigma_i^2=\alpha_i \beta_i$ for $0 < \alpha_i < \beta_i < +\infty$. Then, up to the constant $C$, maximizing $\mathcal{J}(\Theta; \sigma_1, ..., \sigma_n)$ over $\Theta$ is equivalent to maximizing
\begin{align}
& \quad\sum_{i=1}^{n}\left(-\frac{z_{i}^{2}}{2 \alpha_{i} \beta_{i}}+ \underbrace{1-\frac{\beta_{i} \log \beta_{i}-\alpha_{i} \log \alpha_{i}}{\beta_{i}-\alpha_{i}}}_{\text{a constant $C$ wrt }\Theta} -\log\sqrt{2 \pi}\right)+\log \left|\operatorname{det}J_{\mathbf{g}}\right| \label{YY}\\
&=\sum_{i=1}^{n}\left(-\frac{z_{i}^{2}}{2\left(\beta_{i}-\alpha_{i}\right)}\left(\frac{1}{\alpha_{i}}-\frac{1}{\beta_{i}}\right)-\left.\frac{1}{\sqrt{2 \pi}\left(\beta_{i}-\alpha_{i}\right)}\left(\sqrt{2 \pi} {\Tilde\sigma}_{i} \log \left(\sqrt{2 \pi} {\Tilde\sigma}_{i}\right)-\sqrt{2 \pi} {\Tilde\sigma}_{i}\right)\right|_{\alpha_{i}} ^{\beta_{i}}\right)+\log \left|\operatorname{det}J_{\mathbf{g}}\right| \\
&=\sum_{i=1}^{n}\left(-\frac{z_{i}^{2}}{2} \int_{\alpha_{i}}^{\beta_{i}} \frac{1}{\beta_{i}-\alpha_{i}} \frac{1}{{\Tilde\sigma}_{i}^{2}} d {\Tilde\sigma}_{i}-\frac{1}{\beta_{i}-\alpha_{i}} \int_{\alpha_{i}}^{\beta_{i}} \log \left(\sqrt{2 \pi} {\Tilde\sigma}_{i}\right) d {\Tilde\sigma}_{i}\right)+\log \left|\operatorname{det}J_{\mathbf{g}}\right| \\
&=\sum_{i=1}^{n} \int_{\alpha_{i}}^{\beta_{i}} \frac{1}{\beta_{i}-\alpha_{i}}\left(-\frac{z_{i}^{2}}{2 {\Tilde\sigma}_{i}^{2}}-\log \sqrt{2 \pi} {\Tilde\sigma}_{i}\right) d {\Tilde\sigma}_{i}+\log \left|\operatorname{det}J_{\mathbf{g}}\right| \\
&=\sum_{i=1}^{n} \mathbb{E}_{{\Tilde\sigma}_{i} \sim \mathcal{U}{\left[\alpha_{i}, \beta_{i}\right]}}\left[-\frac{z_{i}^{2}}{2 {\Tilde\sigma}_{i}^{2}}-\log \left(\sqrt{2 \pi} {\Tilde\sigma}_{i}\right)\right]+\log \left|\operatorname{det}J_{\mathbf{g}}\right| \\
&=\sum_{i=1}^{n} \mathbb{E}_{{\Tilde\sigma}_{i} \sim \mathcal{U}[\alpha_{i}, \beta_{i}]}\left[\log \left(\frac{1}{\sqrt{2 \pi} {\Tilde\sigma}_{i}} \exp \left(-\frac{z_{i}^{2}}{2 {\Tilde\sigma}_{i}^{2}}\right)\right)\right] +\log \left|\operatorname{det}J_{\mathbf{g}}\right|\\
&\le \sum_{i=1}^{n} \log \left(\mathbb{E}_{{\Tilde\sigma}_{i} \sim \mathcal{U}\left[\alpha_{i}, \beta_{i}\right]}\left[\frac{1}{\sqrt{2 \pi} {\Tilde\sigma}_{i}} \exp \left(-\frac{z_{i}^{2}}{2 {\Tilde\sigma}_{i}^{2}}\right)\right]\right) +\log \left|\operatorname{det}J_{\mathbf{g}}\right| \\
&=\sum_{i=1}^{n} \log \int_{\alpha_{i}}^{\beta_{i}} \frac{1}{\beta_{i}-\alpha_{i}} \frac{1}{\sqrt{2 \pi} {\Tilde\sigma}_{i}} \exp \left(-\frac{z_{i}^{2}}{2 {\Tilde\sigma}_{i}^{2}}\right) d {\Tilde\sigma}_{i} +\log \left|\operatorname{det}J_{\mathbf{g}}\right| \\
&=\log \prod_{i=1}^{n} \int_{\alpha_{i}}^{\beta_{i}} p\left({\Tilde\sigma}_{i} | \alpha_{i}, \beta_{i}\right) p\left(z_{i} | {\Tilde\sigma}_{i}\right) d {\Tilde\sigma}_{i} +\log \left|\operatorname{det}J_{\mathbf{g}}\right| \\
&= \log \int_{\mathbb{R}_{+}^n} p(\mathbf{z}|\Tilde\sigma_1, ..., \Tilde\sigma_n) \prod_{i=1}^n {p(\Tilde\sigma_i | \alpha_i, \beta_i)} d\Tilde\sigma_1 d\Tilde\sigma_2 \cdots d\Tilde\sigma_n +\log \left|\operatorname{det}J_{\mathbf{g}}\right|
\end{align}
where $J_{\mathbf{g}}:=\partial \mathbf{g}_{\Theta} / {\partial \mathbf{x}}$. Eq.~(A.8) is obtained by applying the Newton-Leibniz formula. Eq.~(A.12) is obtained by Jensen's inequality. Eq.~(A.15) is obtained from the property that an integral over the support of the uniform distribution $\mathcal{U}[\alpha_i, \beta_i]$ is equal to an integral over an interval covering the support $[\alpha_i, \beta_i] \subset \mathbb{R}_{+}$ where $0 < \alpha_i < \beta_i < +\infty$ for all $i$'s.

Note that the constant $C$ can be rewritten into
\begin{equation}
C = \sum_{i=1}^n{\frac{\beta_{i}\left(1-\log \beta_{i}\right)-\alpha_{i}\left(1-\log \alpha_{i}\right)}{\beta_{i}-\alpha_{i}}}.
\end{equation}
Consider a univariate function $f(x):=x(1-\log x)$. It can be readily shown that $f(x)$ is an increasing function within the interval $(0,1]$. It follows that $C>0$, which concludes the proof.
\end{proof}

\section{Training Neural-PCA with Gradient-Stoppping Trick}
To train Neural-PCA, one needs to evaluate the log-determinant of Jacobian in the objective (3), which can be written into a summation of three terms:
\begin{equation}
\log \left|\operatorname{det} \left(\frac{\partial \mathbf{g}_{\Theta}}{\partial \mathbf{x}}\right) \right| = \log \left|\operatorname{det}\left(\frac{\partial \mathbf{h}}{\partial \mathbf{x}} \right)\right| + \log \left|\operatorname{det}\left(\frac{\partial \Phi}{\partial \mathbf{h}}\right)\right| + \log \left|\operatorname{det}\left(\frac{\partial \Psi}{\partial \Phi}\right)\right|
\end{equation}
where the first term can be readily obtained from a regular NF, but the evaluation of the second and the third turns out to be quite expensive. \cite{ionescu2015matrix} derived the SVD variations for a general loss function. However, the optimization of our objective requires higher orders of SVD variations, which complicates the back-propagation process. For the sake of simplicity, we apply the gradient-stopping trick to the last two terms \cite{he2020momentum}. We find that this simple treatment not only largely reduces the computational overheads, but also empirically results in satisfactory performance improvements.

\section{Proof of Theorem 2}
\begin{proof}
The orthogonal projection of $\overline{V}$ onto $SO(n)$ is
\begin{equation}
\begin{split}
\Pi(\overline{V}) 
&=\underset{V \in S O(n)}{\arg \min }\|\overline{V}-V\|_{F} =\underset{V \in S O(n)}{\arg \min }\|\overline{V}-V\|_{F}^{2} \\
&=\underset{V \in S O(n)}{\arg \min }-\frac{2}{M} \operatorname{tr}\left(\sum_{i=1}^{M} {V^{(i)}}^{T} V\right) \\
&=\underset{V \in S O(n)}{\arg \max } \operatorname{tr}\left(\overline{V}^{T} V\right)
\end{split}
\end{equation}
On the other hand, the mean rotation matrix $\mathfrak{M}(V^{(1)}, \cdots, V^{(M)})$ of the orthogonal matrices $V^{(1)}, \cdots, V^{(M)}$ is given by
\begin{equation}
 \begin{split}
\mathfrak{M}\left(V^{(1)}, \cdots, V^{(M)}\right) \\ =\underset{V \in S O(n)}{\arg \min } \sum_{m=1}^{M}\left\|V^{(m)}-V\right\|_{F}^{2}
= \underset{V \in S O(n)}{\arg \max } \operatorname{tr}\left(\overline{V}^{T} V\right)
 \end{split}
\end{equation}
Therefore, $\mathfrak{M}\left(V^{(1)}, \cdots, V^{(M)}\right) = \Pi(\overline{V})$. Hence, given the arithmetic mean $\overline{V}$, the mean rotation can be found by solving the optimization problem $\Tilde{V} = {\arg \max }_{V \in SO(n)} \operatorname{tr}(\overline{V}^TV)$ as discussed in the main paper.

\end{proof}

\section{The Close-Form Solution to The Optimization Objective: $\Tilde{V} = {\arg \min }_{V \in SO(n)} \|\overline{V}-V\|_{F}$}

First, we show $\Tilde{V} = {\arg \min }_{V \in SO(n)} \|\overline{V}-V\|_{F} = {\arg\max}_{V \in SO(n)}\operatorname{tr}(\overline{V}^TV)$:

\begin{align}
\lVert \overline{V} - V \rVert_F^2 
= \operatorname{tr}((\overline{V}-V)^T(\overline{V}-V)) \\
= \operatorname{tr}(\underbrace{\overline{V}^T\overline{V}}_{\text{a constant wrt.} V} - 2\overline{V}^TV + \underbrace{V^TV}_{=I})
\end{align}
where $\overline{V}$ is the arithmetic mean, i.e. $\overline{V}=\sum_{i=1}^{M}{V^{(i)}/M}$, which is a constant with regard to $V$. Therefore, the objective is boiled down to finding a matrix $V$ that maximizes $\operatorname{tr}(\overline{V}^TV)$. 

Second, we take SVD of $\overline{V}^T$, which yields $\overline{V}^T = P \Lambda Q^T$. Then the objective becomes $\text{tr}(\overline{V}^TV) = \text{tr}(P \Lambda Q^TV) = \text{tr}(\Lambda Q^TVP)$. Let $R = Q^TVP$. Note that $R$ is an orthogonal matrix and it follows that each element in $R$ is less than or equal to $1$, i.e. $R_{ij}\le 1$, for all $i, j$. Therefore, the objective has the upperbound,
\begin{equation}
\operatorname{tr}(\Lambda R) = R_{11} \lambda_{1} + ... + R_{nn} \lambda_{n} \le \lambda_{1} + ... + \lambda_{n}.
\end{equation}
where $\lambda_{1}, ..., \lambda_{n}$ are the diagnal entries of the matrix $\Lambda$.

The equality holds true iff $V=Q P^T$, which yields the optimal solution of $\Tilde{V} = {\arg \min }_{V \in SO(n)} \|\overline{V}-V\|_{F}$.

\section{Detailed Analysis of Toy Dataset Results}

Comparison among Neural-PCA, Baseline-NIG, Baseline-BN and Neural-PCA-IG indicates that the PCA block and the non-isotropy of base density are two essential building blocks in capturing principal components. Comparison with Baseline-R and Baseline-BN-R further suggests that Neural-PCA surpasses the trainable rotation layer in capturing major discriminative features. Though theoretically a trainable rotation layer admits universal approximation to any rotation matrix in $SO(n)$, in practice it is hard for the flow to capture the ``PCA-rotation" without an explicit inductive bias due to its agnostics of data variation. The optimization objective easily converges to a local minimum that corresponds to non-PCA-rotation: as shown in Figure~\ref{fig:twospirals2d}, the rotation angles learned by Baseline-R and Baseline-BN-R are almost zeros whereas Neural-PCA achieves non-trivial rotation of roughly $-\pi/4$ that results in clearer separability.

\section{Comparison with Post-PCA models}

\setlength{\tabcolsep}{4pt}
\begin{table}[b]
\small
\begin{center}
\caption{\small Ablation study on Post-PCA models in terms of classification accuracy.}
\setlength{\tabcolsep}{3.0mm}
\begin{tabular}{l|cc}
\toprule[1.5pt]
Models & MNIST & FashionMNIST  \\
\hline
Baseline  &  $0.6908$ &  $0.5997$ \\
+ Post-PCA  &   $0.7123$ &  $0.5956$ \\
\hline
Baseline-NIG  &  $0.6066$ &  $0.6114$ \\
+ Post-PCA  &   $0.6521$ &  $0.6418$ \\
\hline
NeuralPCA  &  $0.8192$ &  $0.7188$ \\
+ Post-PCA  &   $0.8173$ &  $0.6894$ \\
\hline
NeuralPCA-IG  &  $0.7223$ &  $0.6582$ \\
+ Post-PCA  &   $0.7162$ &  $0.5986$ \\
\hline
Baseline-BN  &  $0.6307$ & $0.5299$ \\
+ Post-PCA  &   $0.7150$ & $0.6399$ \\
\hline
Baseline-R  &  $0.1028$ & $0.0960$ \\
+ Post-PCA  &   $0.1077$ & $0.1010$ \\
\hline
Baseline-BN-R  &  $0.1001$ & $0.1003$ \\
+ Post-PCA  &   $0.1007$ &  $0.1050$ \\
\hline
\bottomrule[1.5pt]
\end{tabular}
\end{center}
\end{table}

\begin{figure}[h]
\centering
\includegraphics[width=\textwidth]{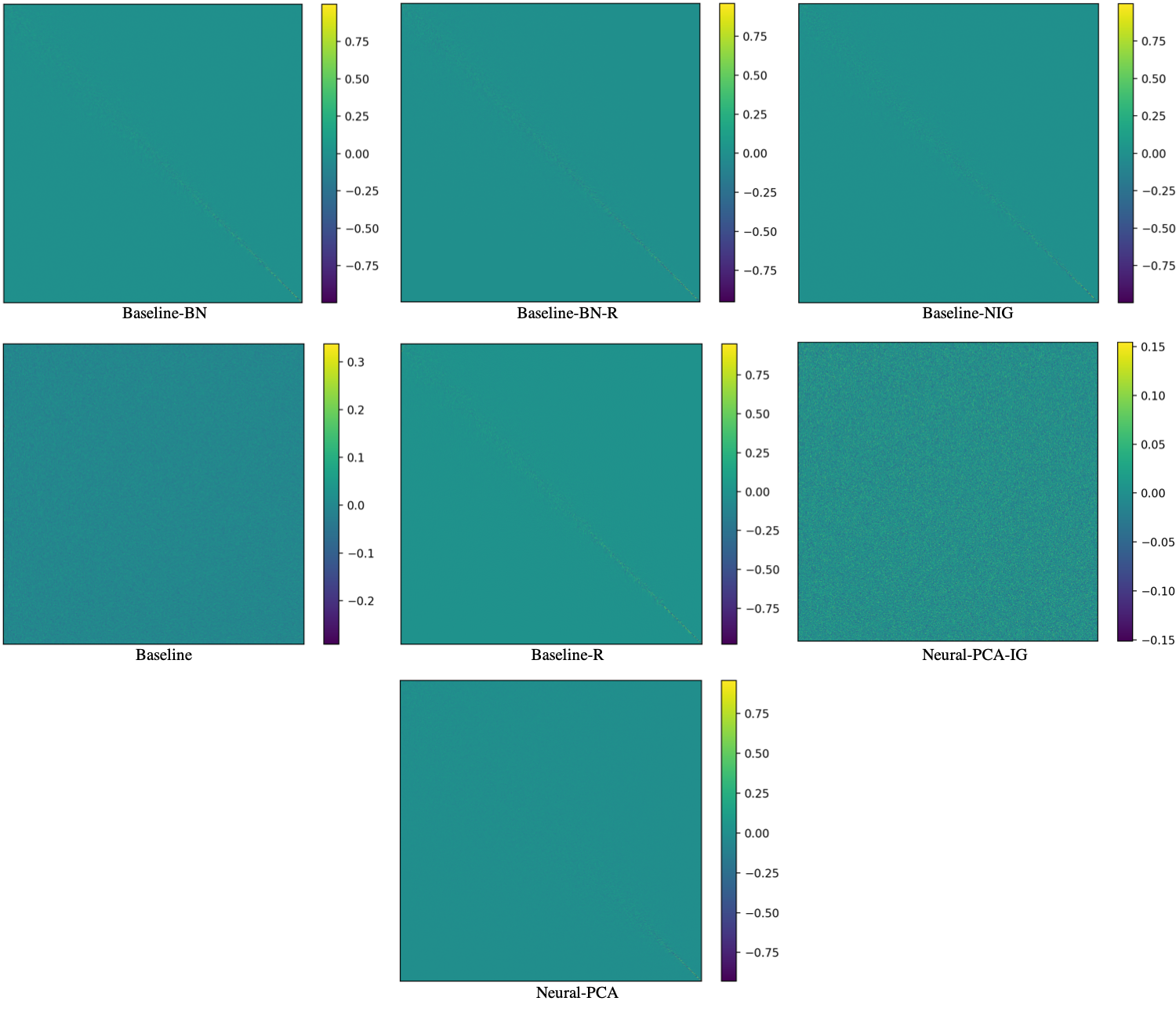}
\caption{\small The rotation matrices $V$ fitted from Post-PCA models, respectively (zoom in for closer scrutiny).}
\label{fig:postpca_v}
\end{figure}

This section demonstrates the necessity of training flow with the proposed PCA-Block jointly in an end-to-end manner. As in the synthetic data setting, we consider post-PCA models based on baseline features learned by Baseline, Baseline-NIG, Baseline-BN, Baseline-BN-R and Neural-PCA-IG. From Table A.1, we observe little differences in accuracy between baseline models and their Post-PCA counterparts. Figure~\ref{fig:postpca_v} demonstrates the rotation matrix $V$ fitted from these Post-PCA models, respectively, which verifies these little differences: expect for IG (isotropic Gaussian) models including Baseline and Neural-PCA-IG, the rotation matrices fitted from other models all exhibit diagonal structures (close to identity matrix with zero rotation angle). This suggests that the representations learned by these models have already been axis-aligned as seen in the synthetic dataset. And therefore Post-PCA on them would not result in non-trivial rotation further. This justifies the necessity of training flow with PCA-Block end-to-end jointly.

\clearpage
\section{Generation Quality and Density Estimation}

\begin{figure*}[t]
\centering
\includegraphics[width=1\textwidth]{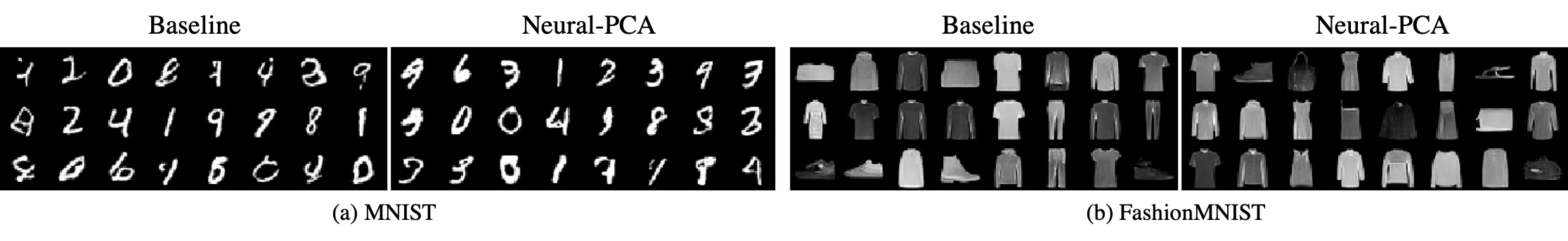}
\caption{\footnotesize Uncurated samples drawn from Baseline and Neural-PCA on two real datasets.}
\label{fig:image_quality}
\end{figure*}

\begin{table}
\centering
\small
\caption{Averaged test negative log-likelihood in bits per dimension (BPD, the lower the better).}\label{table:test_bpd}
\begin{tabular}{c|cc}
\toprule[1.5pt]
& MNIST & FashionMNIST \\
\midrule
RealNVP \cite{dinh2016density} & $1.06$ & $3.12$ \\
Glow \cite{kingma2018glow} &  $1.05$   &  $2.96$  \\
MAF \cite{papamakarios2017masked} &  $1.89$ & $-$ \\
SOS \cite{jaini2019sum} &  $1.81$   & $-$  \\
RQ-NSF(C) \cite{durkan2019neural}  &  $0.99$   &  $2.85$ \\
\midrule
Neural-PCA (built on RQ-NSF(C)) &  $1.02$   &  $2.89$ \\
\bottomrule[1.5pt]
\end{tabular}
\end{table}
\setlength\intextsep{-0.5ex}

We show that Neural-PCA exhibits competitive negative log-likelihood on test splits of image datasets as compared with its baseline. As shown in Table~\ref{table:test_bpd}, the PCA block we introduce into a regular normalizing flow does not compromise the capability of density estimation. 

In terms of sample quality, Neural-PCA can also generate visually authentic images as Baseline. By solving $\Tilde{V} = {\arg \min }_{V \in SO(n)} \|\overline{V}-V\|_{F}$ for the mean rotation of matrices in $SO(n)$, the PCA layer can be properly inverted by simply post-multiplying the transpose of the mean rotation matrix. Figure~\ref{fig:image_quality} demonstrates uncurated samples drawn from Baseline and Neural-PCA.

\clearpage
\section{Mean Rotation Analysis.} 
\begin{figure}[t]
\centering
\includegraphics[width=0.8\textwidth]{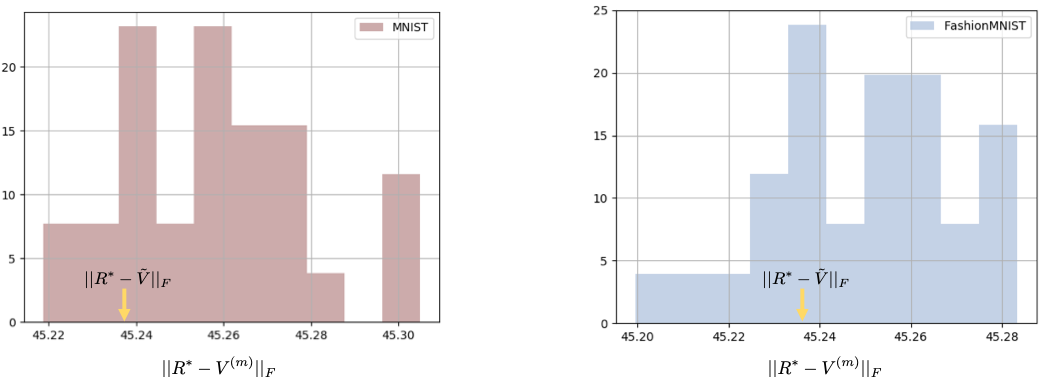}
\caption{\small The empirical distribution of rotation matrix distances. Here, $V^{(m)}$ is determined by SVD of PCA layer for the $m$th batch, $m=1,...,M$. $\Tilde{V}$ is the mean rotation given by $\Tilde{V} = {\arg \min }_{V \in SO(n)} \|\overline{V}-V\|_{F}$ in the main text. $R^*$ is the optimal rotation learned by Baseline-BN-R (or Baseline-R). }
\label{fig:fnorm_dist}
\end{figure}

Figure~\ref{fig:fnorm_dist} demonstrates the empirical distribution of rotation matrix distances, $||R^* - V^{(m)}||_F$ for all batches $m=1, ..., M$. It is observed that distances between the rotation matrix $R^*$ learned by Baseline-BN-R (or Baseline-R) and the mean rotation $\Tilde{V}$ by the PCA layer are conspicuous. This is in line with the results shown in the setting of Two-Spiral in which a trainable rotation layer of Baseline-BN-R (or Baseline-R) learns rotation angles of nearly zero whereas PCA layer yields nontrivial variation-aware rotations leading to better representation.

Figure~\ref{fig:fnorm_dist} also suggests a small amount of variances on the rotation matrix $V^{(m)}$ thanks to a large batch size (e.g. 1800). The matrix difference ranges from $45.22$ to $45.30$ on MNIST and from $45.20$ to $45.28$ on FashionMNIST, respectively. This suggests the limitation of Neural-PCA: one has to use a large batch size during training in order to make sure PCA and the mean rotation matrix $\Tilde{V}$ can be approximated accurately. Future work might consider approximation even with a small batch size.

\section{Latent Interpolation}
We show that Neural-PCA also allows for better interpretability. By compressing the information in the data into a fewer dimensions (in a lossy manner), the data becomes easier to interpret: one can interpret each latent dimension by varying that dimension individually and inspect how the generated images change accordingly. This section demonstrates how images vary as interpolation is performed along a few leading or trailing latent dimensions, simultaneously. Specifically, we consider the following linear interpolation:
\begin{equation}
\begin{split}
\mathbf{z} = \biggl(\underbrace{-2\lambda+2(1-\lambda), \cdots, -2\lambda+2(1-\lambda)}_{128\text{ dims}}, \underbrace{0, \cdots, 0}_{(n-128)\text{ dims}}\biggr) \\
\quad \text{for interpolation along latent leading dimensions;}
\end{split}
\end{equation}

\begin{equation}
\begin{split}
\mathbf{z} = \biggl(\underbrace{0, \cdots, 0}_{(n-128)\text{ dims}}, \underbrace{-2\lambda+2(1-\lambda), \cdots, -2\lambda+2(1-\lambda)}_{128\text{ dims}}\biggr) \\
\quad \text{for interpolation along latent trailing dimensions;}
\end{split}
\end{equation}
where $\lambda$ goes from $1$ down to $0$.

\begin{figure*}[t]
\centering
\includegraphics[width=\textwidth]{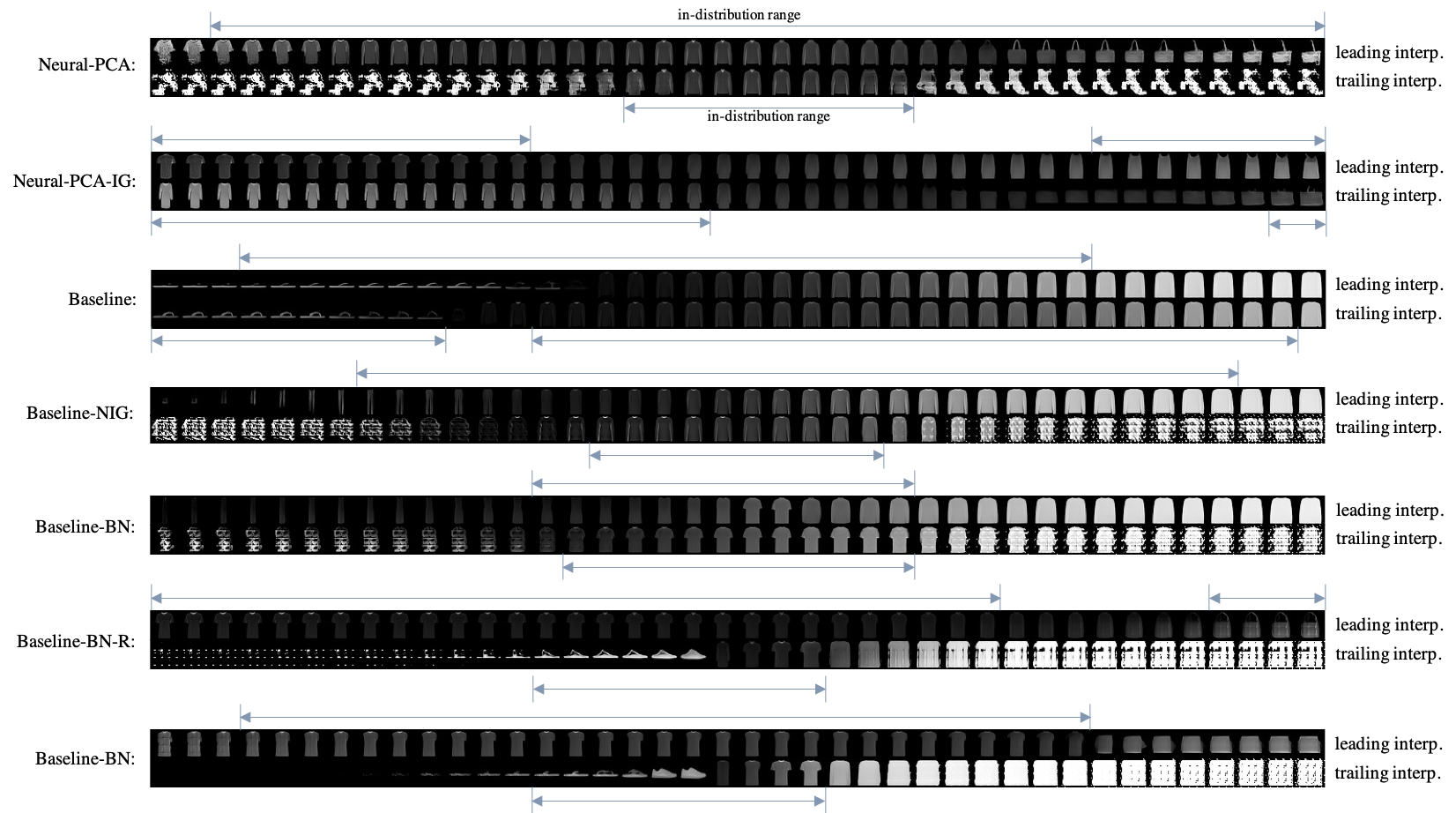}
\caption{\footnotesize Latent interpolation for all models in question. For each model, the first row showcases the resultant images by interpolating along the first 128 latent dimensions with the rest fixed to zeros; the second row showcases those by interpolating along the last 128 latent dimensions with the rest fixed to zeros. Interpolation range is from $-2$ to $2$. In-distribution ranges are marked by arrows.}
\label{fig:latent_walk}
\end{figure*}

Figure~\ref{fig:latent_walk} showcases the interpolation results for all models in question. Qualitative comparison suggests that when interpolated along the leading dimensions, Neural-PCA allows for a longer in-distribution range, yielding images (that are in-distribution) with more variations and better visual fidelity, whereas interpolation along the trailing dimensions results in anomaly noisy samples (that are out-of-distribution) and unvarying images. This corroborates that Neural-PCA tends to store most representative information in leading dimensions and negligible information in trailing dimensions. Though other models with non-isotropic Gaussians (Baseline-NIG, Baseline-BN, Baseline-BN-R, Baseline-BN) also manifest relatively longer in-distribution ranges in leading interpolation than in trailing interpolation, they suffer from mode collapse, generating unvarying images of inconspicuous differences. 
Without explicit inductive biases (e.g. Neural-PCA), flow-based models are agnostic in finding good rotation that leads to better representation. 

\section{Experimental Details}
For reproducing the results reported in the paper, this section details experimental settings. The common hyperparameters of all models listed in Table 1 are the same unless otherwise stated. The default configurations of the baseline models (FFJORD and RQNSF(C)) are used as specified in \cite{grathwohl2018ffjord} and in Appendix B.3 of \cite{durkan2019neural}, respectively. Neural-PCA is built upon these baseline models by adding the PCA block at the end. Code for Baseline implementation can be found \texttt{https://github.com/rtqichen/ffjord} and \texttt{https://github.com/bayesiains/nsf}.

Each experiment is carried out using one Tesla V100 32GB GPU. It takes roughly 56 hours to train Neural-PCA on realistic image datasets and 8 hours on the toy dataset. The number of training iterations on the Two-Spiral dataset is 10,000 and 100,000 on realistic image datasets. During training, we use held-out data (validation splits of datasets) to select the best model checkpoint for evaluations of density estimation, sample quality and downtream metrics on the test splits.

\end{document}